\documentclass[lettersize,journal]{IEEEtran}
\usepackage{amsmath,amsfonts}
\usepackage{algorithmic}
\usepackage{algorithm}
\usepackage{array}
\usepackage[caption=false,font=normalsize,labelfont=sf,textfont=sf]{subfig}
\usepackage{textcomp}
\usepackage{orcidlink}
\usepackage{stfloats}
\usepackage{url}
\usepackage{verbatim}
\usepackage{graphicx}
\usepackage{cite}
\usepackage{booktabs}
\usepackage{multirow}
\usepackage{pifont}
\newcommand{\cmark}{\ding{51}}
\newcommand{\xmark}{\ding{55}}
\def\ourmethod{KiMoI}
\def\ourmethodfull{Kinematic Model for facial motion Inconsistencies}

\begin{document}

\title{Beyond Flicker: Detecting Kinematic Inconsistencies for Generalizable Deepfake Video Detection} 

\author{\IEEEauthorblockN{Alejandro Cobo\orcidlink{0009-0007-7967-6837}\IEEEauthorrefmark{1}}
\and
\IEEEauthorblockN{Roberto
Valle\orcidlink{0000-0003-1423-1478
}\IEEEauthorrefmark{1}}
\and
\IEEEauthorblockN{José M. Buenaposada\orcidlink{0000-0002-4308-9653}\IEEEauthorrefmark{2}}
\and
\IEEEauthorblockN{Luis Baumela\orcidlink{0000-0001-6910-4359}\IEEEauthorrefmark{1}}}

\maketitle

\begingroup\renewcommand\thefootnote{}
\footnotetext{\IEEEauthorrefmark{1} Affiliation with Departamento de Inteligencia Artificial, Universidad Politécnica de Madrid, Madrid, Spain. \{alejandro.cobo,roberto.valle,luis.baumela\}@upm.es}
\footnotetext{\IEEEauthorrefmark{2} Affiliation with Departamento de Informática y Estadística, Universidad Rey Juan Carlos, Móstoles, Spain. josemiguel.buenaposada@urjc.es}
\endgroup

\begin{abstract}
    Generalizing deepfake detection to unseen manipulations remains a key challenge. A popular approach to tackle this issue is to train a network with pristine face images that have been manipulated with hand-crafted artifacts to extract more generalizable clues. While effective for static images, extending this to the video domain still remains an open issue. Existing methods model temporal artifacts as frame-to-frame instabilities, overlooking a key vulnerability: the violation of natural motion dependencies between different facial regions.
    In this paper, we propose a synthetic video generation method that creates training data with subtle kinematic inconsistencies. We train an autoencoder to decompose facial landmark configurations into motion bases. By manipulating these bases, we selectively break the natural correlations in facial movements and introduce these artifacts into pristine videos via face morphing. A network trained on our data learns to spot these sophisticated biomechanical flaws, achieving state-of-the-art generalization results on several popular benchmarks.
\end{abstract}

%-------------------------------------------------------------------------
\section{Introduction}

\textit{Deepfakes} refer to audiovisual media of human subjects whose content has been manipulated or entirely generated by artificial methods. Although these techniques are useful in scenarios such as filmmaking or social media applications, they can also be exploited for malicious purposes, including spreading misinformation and identity theft.

Consequently, as deepfake generation methods grow in sophistication, so does the need for advanced detection systems that can reliably identify manipulated media, particularly from unseen generation methods. A successful approach to improve generalization is the use of pseudo-fake generation techniques~\cite{Li20a, Chen22, Shiohara22, Wang23a, Larue23, Cai25, Nguyen25, Yan25b}, in which fake samples are generated from pristine data to train a predictive model. This allows for a higher controllability of the data presented to the model, including samples that are very similar to their pristine counterparts, which forces the network to extract more informative representations to distinguish between real and manipulated faces.

\begin{figure}[!t]
    \centering
    \includegraphics[width=0.47\textwidth]{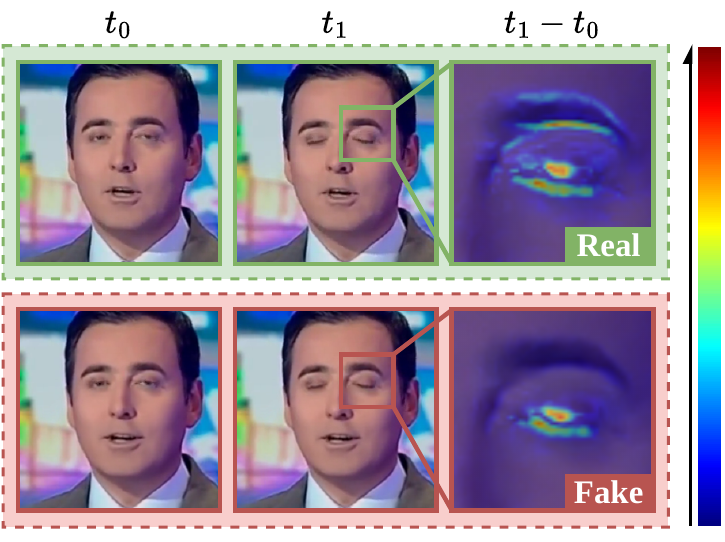}
    \caption{Example of temporal artifacts introduced by a deepfake generation method~\cite{Thies19}. It fails to accurately replicate the correlation between eyebrow and eyelid movements observed during eye closure in the original video.}
    \label{fig:deepfake-artifacts}
\end{figure}

Pseudo-fake deepfake detection approaches mainly focus on detecting \textit{spatial artifacts} such as blending boundaries and texture inconsistencies~\cite{Shiohara22, Li20a} and \textit{temporal artifacts} associated with the digital fingerprint of different video sources~\cite{Wang23a} and the flicker or drift of individual facial features over time~\cite{Cai25, Nguyen25, Yan25b}. However, they do not investigate the violation of natural motion dependencies between different facial parts. Although some recent detection methods rely on hand-crafted geometric rules to spot muscle inconsistencies~\cite{Liao23}, data-driven synthesis of such complex temporal artifacts remains unexplored. For example, the fake in Fig.~\ref{fig:deepfake-artifacts} would go unnoticed since there is no temporal drift, but a very subtle uncorrelation between the movement of the eyebrow and the eyelid. We argue that modern deepfake generation methods are progressively introducing less obvious temporal artifacts, and present pseudo-fake approaches could be rendered insufficient.

\begin{figure*}[!t]
    \centering
    \includegraphics[width=0.8\textwidth]{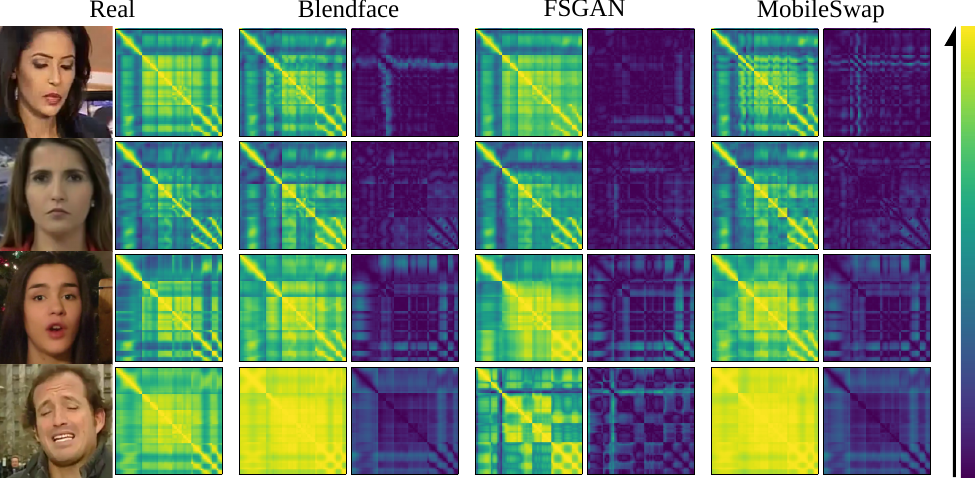}
    \caption{Differences in landmark motion correlation (see Eq.~\ref{eq:facial-movement}) between real videos and several deepfake generation methods included in the DF40 dataset~\cite{Yan23a}. Each row shows correlations for a different video sequence. Next to each correlation heatmap for fake videos, a difference with the correlation heatmap of the real video is shown.}
    \label{fig:real-fake-corr-diff}
\end{figure*}

Real human faces are not collections of independent parts; they are complex systems governed by an underlying musculature and skeletal structure. The movement of one part is biomechanically linked to that of others, creating a set of natural correlated motion patterns. 
Deepfake models, especially those that might learn to render different facial parts in a semi-independent manner, can fail to reproduce these complex correlations (see Fig.~\ref{fig:real-fake-corr-diff}). This results in motions without temporal artifacts that look plausible in isolation but are subtly wrong in their relationship to each other.

Hence, in this paper, we propose a new data-driven methodology to synthesize subtle temporal artifacts. Our approach is built on a novel generative model of facial kinematics, which we use within a flexible synthesis framework to create a diverse range of subtle motion inconsistencies. This allows us to train detectors that are more sensitive to the sophisticated forgeries produced by modern deepfake models.

In summary, our contributions can be listed as follows:
\begin{enumerate}
    \item A data-driven generator for facial motion artifacts. We propose a generative model, based on an autoencoder architecture, that learns a structured representation of facial kinematics.
    \item A flexible synthesis pipeline for non-rigid temporal forgery.
    We introduce a synthesis process that adapts established face warping techniques for a new purpose: applying targeted, non-rigid temporal artifacts driven by our learned motion generator. 
    This approach goes beyond previous methods that have focused on more holistic, face-wide transformations, allowing for a new level of detail and realism in the synthesized artifacts.
    \item A new state-of-the-art in generalizable deepfake video detection. We empirically demonstrate that a hybrid training strategy, combining existing spatial pseudo-fakes with our proposed temporal artifacts outperforms previous approaches on multiple leading benchmarks.
\end{enumerate}

%-------------------------------------------------------------------------
\section{Related work}
\label{sec:related-work}

\textbf{Deepfake detection}. Early works in deepfake detection focused on extracting image-level clues, either in the RGB space~\cite{Chai20, Carvalho13, Carvalho15, Rossler19} or the frequency space~\cite{Stuchi17, Durall19, Qian20, Wang20a, Jeong22, Miao22}, but do not generalize well to unseen manipulation techniques. Some methods leverage video inputs to take advantage of more informative spatio-temporal features~\cite{Cobo25, Haliassos21, Haliassos22, Wang23a, Xu23, Zheng21, Hu22, Pang23, Wang23b, Yu23, Zhao23, Guo25, Kim25}. A popular approach among these methods includes the detection of temporal artifacts with different window sizes to capture local and global inconsistencies~\cite{Pang23, Yu23}. Wang et al.~\cite{Wang23b} explore the inconsistency between the movement of the entire face and the mouth region in deepfake videos.

Moreover, since spatial artifacts usually dominate deepfake videos, it is not easy for the model to correctly learn to identify temporal clues~\cite{Zheng21, Wang23a}. To mitigate this issue, some techniques have been proposed. One example is FTCN~\cite{Zheng21}, where the layers of a 3D ResNet are adapted to process information in the temporal direction only, and long-range temporal dependencies are modeled by transformer layers. Furthermore, AltFreezing~\cite{Wang23a} proposed a novel training scheme, where spatial and temporal convolutional kernels are learned intermittently. Spatio-temporal information can also be used to track biometric and kinematic information to detect inconsistencies in fake videos, such as head pose~\cite{Yang19}, facial landmarks~\cite{Wang20b, Li21, Sun21, Liao23} and heart rate~\cite{Ciftci20a, Ciftci20b, Hernandez20}. Among these methods, the most similar to our approach is FAMM~\cite{Liao23}, which trains a GRU on hand-crafted features from facial landmarks to model facial muscle interactions. However, the simplification of video data into geometric features to train the classifier causes significant information loss, as the model cannot capture texture data and the landmark annotations can contain noise introduced by the face alignment network.

To improve generalization towards different deepfake generation methods, many works leverage techniques such as contrastive learning~\cite{Guo23, Larue23}, attention mechanisms~\cite{Zhao21, Yang21} or self-supervision~\cite{Chen22, Haliassos22, Luo23, Wang25}. Haliassos et al.~\cite{Haliassos21, Haliassos22} focus on modeling the movement of talking faces by pre-training on real-face datasets, and leveraging the learned features to guide the deepfake detection network. However, the fake videos used to train the detector~\cite{Rossler19} contain very specific temporal clues, which hinders its generalization capabilities to unseen manipulation methods. Other methods perform multi-modal audio-visual detection by incorporating information from audio and visual modalities~\cite{Liu24, Oorloff24, Yu24, Smeu25b}.

Several methods focus on identifying extra information related to the manipulation procedure performed on a video, other than the real or fake label. Sequential Deepfake Detection refers to predicting the different sequential steps followed by multi-step manipulation methods~\cite{Shao22, Hong24, Xia24}. Other methods try to identify the particular frames that have been manipulated in deepfake videos that contain a mix of real and fake frames~\cite{Masi20, Li25b}, or the region containing the manipulated pixels within an image~\cite{Liu22}.

Recent approaches focus on adapting large foundational models for the deepfake detection task~\cite{Cai25, Han25, Li25a, Shao25, Smeu25a, Yan25a, Yan25b, Yang25}, the usage of multi-task frameworks that include auxiliary tasks related to deepfake detection~\cite{Cao22, Cobo25, Nguyen25} or networks aided with external information such as semantic regions~\cite{Peng25, Zou25}.

Deepfake-Adapter~\cite{Shao25} leverages global and local information of features extracted from a frozen CLIP model~\cite{Radford21} to improve generalization. FCG~\cite{Han25} proposes an adapter for CLIP features to process spatial and temporal information with a facial component guiding module to focus on key facial features. FakeRadar~\cite{Li25a} applies clustering modeling techniques to CLIP features to identify distributional gaps between real videos, known forgeries and unseen manipulations. Effort~\cite{Yan25a} applies SVD decomposition to the latent space of foundational models and freezes the principal components while training the remaining components, improving generalization.

RECCE~\cite{Cao22} uses decoder features from a denoising autoencoder network trained to reconstruct real face images to train a deepfake detection classifier. SFA~\cite{Cobo25} employs face alignment as an auxiliary task to force the network to model the movement of the face and spot unnatural motions found on deepfake videos. Other methods~\cite{Nguyen19, Li20a, Nguyen24, Nguyen25} train the network to predict heatmaps containing the blending region of face-swap deepfakes.

However, generalization to unseen manipulation methods requires not only an appropriate model architecture, but most importantly, the correct data to train the network, as it is crucial for the model to be guided toward extracting the most relevant features. Thus, we propose a new framework to generate synthetic videos that is independent of any model architecture and can be combined with existing solutions.

\noindent \textbf{Pseudo-fake generation}. A popular approach to improve generalization is to leverage a data augmentation technique termed pseudo-fake generation, where manipulated examples are generated from pristine data as the model trains. By training with real samples slightly modified with artifacts designed to mimic generic clues common to most deepfake generation methods, we can guide the network to extract more general features and avoid overfitting to specific clues introduced by particular deepfake generation methods, which yields better generalization results. This technique also facilitates training on large-scale datasets, as only real videos are needed to train the deepfake detection network.

Early pseudo-fake generation methods simulated face swap by blending two images of the same subject in different scenarios~\cite{Li20a}, and this was later improved by using two slightly different copies of the same image (SBI)~\cite{Shiohara22}. Other methods use self-supervision to generate adversarial fake samples~\cite{Chen22}. These methods work well for detecting face swap–style manipulations, but under-perform when exposed to other forgery types, such as face reenactment, or methods that leave less obvious blending traces (see Tab.~\ref{tab:pseudo-fake-ablation}). Despite recent efforts to perform more subtle and localized pseudo-fake samples to train deepfake detection systems~\cite{Sun25}, the lack of temporal information in these approaches renders these methods to miss a key factor to identify a wider variety of clues in deepfake videos.

This has motivated subsequent research to explore the generalization of this idea to video sequences~\cite{Wang23a, Cai25, Yan25b, Nguyen25}. \textit{Altfreezing} \cite{Wang23a} performs clip-level blending (CBI), where faces in two video clips are blended frame-by-frame. \textit{FakeSTormer}~\cite{Nguyen25} and \textit{DeepShield}~\cite{Cai25} apply the SBI pipeline jointly to all frames in a video clip. \textit{VB}~\cite{Yan25b} applies global affine transformations localized to inner facial regions, and constructs a sequence of independently manipulated frames.
However, in CBI~\cite{Wang23a}, substantial unrealistic artifacts can be introduced if the motion portrayed in the two clips differs greatly. In the case of~\cite{Cai25, Nguyen25, Yan25b}, the complex nature of correlated movements between facial regions is not considered, as this is approximated by a simple face-wide blending procedure. It is also important to note that temporal artifacts in deepfake videos include not only extra movements not present in the original sequence, but also the lack of correlation between the motion of different facial components, as shown in Fig.~\ref{fig:deepfake-artifacts}. Hence, we propose to generate temporal artifacts following a data-driven approach, as these inconsistencies cannot be modeled with analytical approximations.

%-------------------------------------------------------------------------
\section{Method}

\begin{figure*}[!t]
    \centering
    \includegraphics[width=0.65\textwidth]{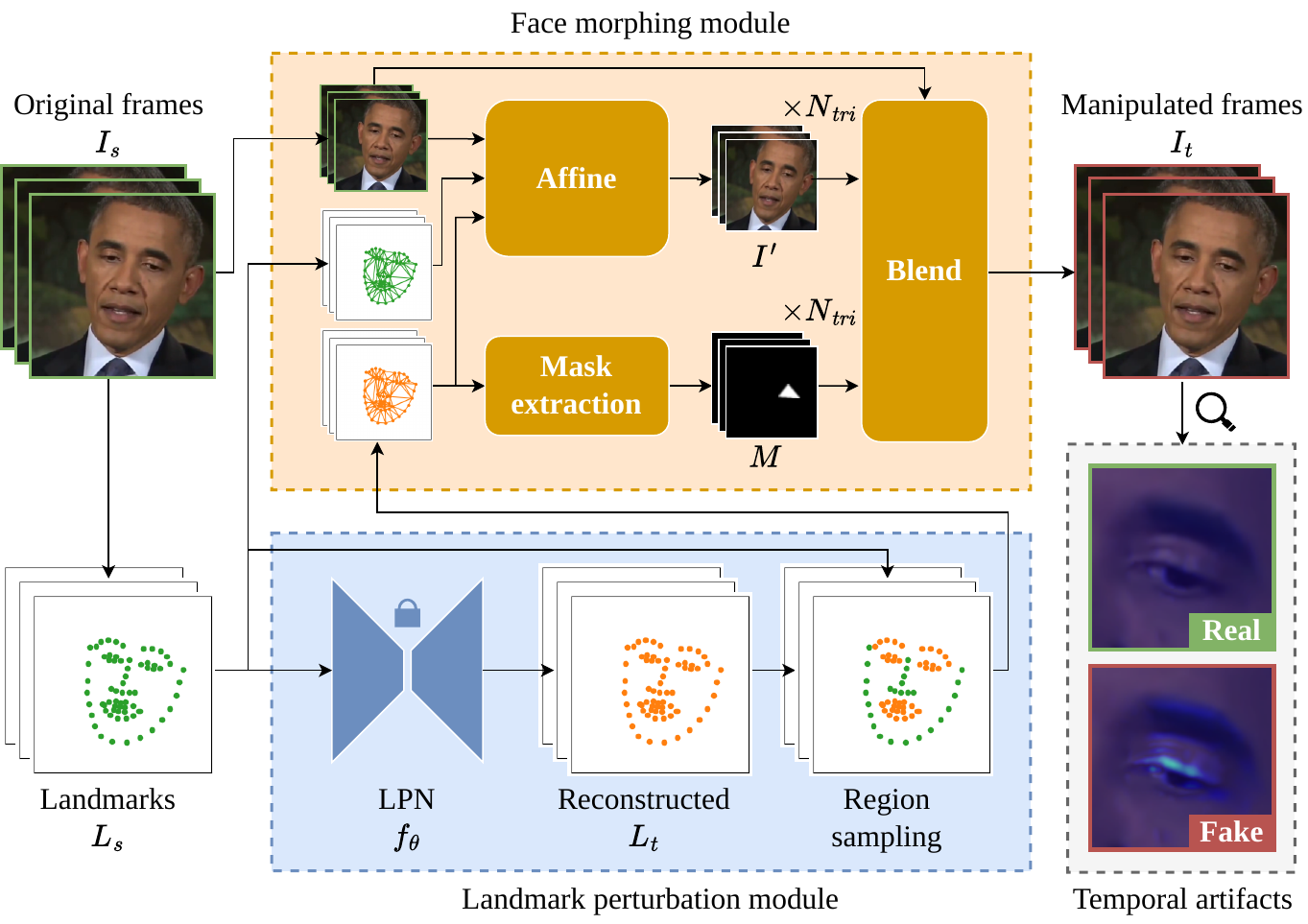}
    \caption{Overview of our method. We leverage a pretrained Landmark Perturbation Network (LPN) that is able to introduce subtle temporal artifacts to landmark sequences extracted from real videos. We then introduce these artifacts to the original frames by modulating facial regions to match the movement of the manipulated landmark sequence. The resulting frames contain generic temporal clues that can be used to train deepfake video detection models.}
    \label{fig:method}
\end{figure*}

Fig.~\ref{fig:method} shows an overview of our pseudo-fake generation framework, called \ourmethodfull{} (\ourmethod{}). First, we extract landmark annotations from a frame sequence and leverage a pretrained Landmark Perturbation Network (LPN) to introduce realistic temporal artifacts to the landmark sequence. This yields a new  sequence that closely resembles the original, but with added subtle differences. Then, we apply a face morphing pipeline to distort facial regions in the original frames to follow the movement portrayed in the reconstructed landmarks, resulting in a pseudo-fake video containing temporal artifacts. In the following sections, we explain in more detail the various parts of our method.

%-------------------------------------------------------------------------
\subsection{Problem formulation}
\label{sec:problem-formulation}

We aim to generate pseudo-fake videos by introducing subtle temporal artifacts to pristine frames sequences. To do so, we first simplify the definition of facial movement to a sequence of $N_{\mathit{lnd}}$ sparse facial landmarks~\cite{Gross08}, $L \in \mathbb{R}^{T \times N_{\mathit{lnd}} \times 2}$, where $T$ is the number of frames in a video clip. We define the facial movement portrayed in the sequence, $\delta L \in \mathbb{R}^{(T-1) \times N_{\mathit{lnd}} \times 2}$, as the difference between landmark coordinates in consecutive frames:
\begin{equation}
    \delta L = \left\{L^{i+1} - L^{i}\right\}_{i=1}^{T-1}.
    \label{eq:facial-movement}
\end{equation}

Therefore, given two aligned landmark sequences, $L_R$ and $L_F$, portraying the same scene in real and fake videos, respectively, we can calculate the temporal artifacts introduced by a manipulation method as:
\begin{equation}
    \Delta_t = \delta L_F - \delta L_R.
    \label{eq:temporal-artifacts}
\end{equation}

Since facial movements involve an extraordinarily complex interaction among different muscles and organs, we hypothesize that $\Delta_t$ cannot be accurately approximated by analytical means, such as global affine transformations, especially considering the rapid development of deepfake generation techniques that may account for temporal consistency. This is also shown in Fig.~\ref{fig:real-fake-corr-diff}, which depicts the difference between correlation heatmaps computed from landmark motions for real videos ($\delta L_R$) and several deepfake generation methods ($\delta L_F$) from the DF40 dataset~\cite{Yan23a}. Since variations in motion correlation are inconsistent between generation methods and depend on the facial motion portrayed on the video, a global analytical approach is rendered infeasible to implement. Therefore, we propose to approximate these temporal artifacts with a neural network, $f_\theta(\cdot)$, as:
\begin{equation}
    \Delta_t \approx \delta f_\theta(L_R) - \delta L_R.
    \label{eq:delta_t_approx}
\end{equation}

%-------------------------------------------------------------------------
\subsection{Landmark Perturbation Network (LPN)}
\label{sec:lpn}

\begin{figure*}[!t]
    \centering
    \includegraphics[width=0.95\textwidth]{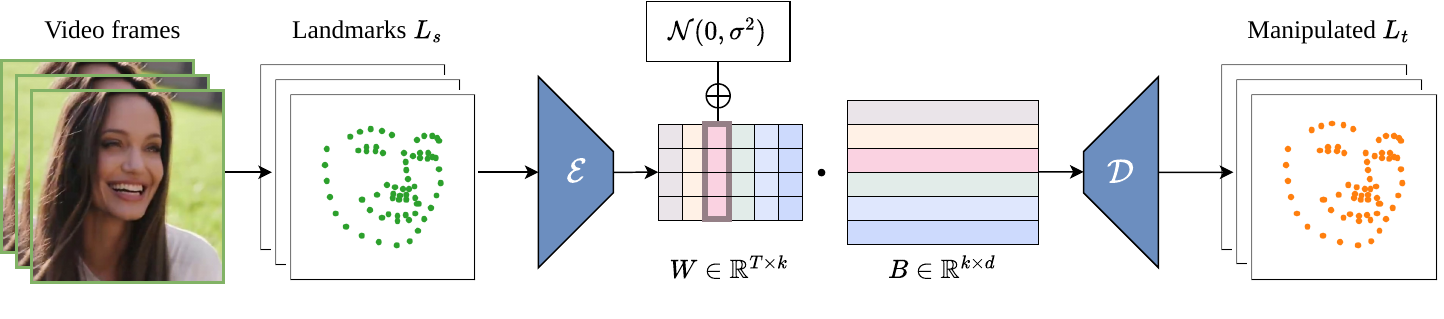}
    \caption{Overview of the Landmark Perturbation Network (LPN). The encoder $\mathcal{E}$ generates a list of weights ($W$) for each time step, and the decoder $\mathcal{D}$ reconstructs the input sequence from a weighted sum of $k$ learnable deformation bases ($B$). In inference time, we generate temporal artifacts by randomly selecting a column in $W$ and adding subtle Gaussian noise to the predicted weights. Since each deformation basis is responsible for different components in the reconstruction of the face, our approach lets us generate a diverse set of semantically meaningful artifacts.}
    \label{fig:lpn}
\end{figure*}

To simulate the temporal artifacts found in deepfake videos, we take advantage of a neural network $f_{\theta}(\cdot)$ trained to reconstruct landmark sequences (see Fig.~\ref{fig:lpn}). Building upon prior methods that utilize face-wide blending operations~\cite{Wang23a, Cai25, Nguyen25, Yan25b}, our work introduces a more flexible approach. We focus on generating a diverse set of localized and non-rigid transformations to provide a richer set of training clues. For this, the network is trained to decompose facial landmark configurations into a series of $k$ learnable deformation bases $B \in \mathbb{R}^{k \times d}$, where $d$ is the size of the latent space. The encoder $\mathcal{E}$ takes the input landmark sequence $L \in \mathbb{R}^{T \times N_{\mathit{lnd}} \times 2}$ and outputs a matrix $W \in \mathbb{R}^{T \times k}$. Each row in $W$ contains the weights for all deformation bases that better approximate the input landmark configuration for a particular time step. The input of the decoder $\mathcal{D}$, $\mathbf{x}_{\mathcal{D}} \in \mathbb{R}^{T \times d}$ is the weighted sum of all deformation bases for each time step $\mathbf{x}_{\mathcal{D}} = W \cdot B$. The decoder then reconstructs the original sequence relying only on this weighted sum of deformation bases. 

In practice, we implement $f_\theta(\cdot)$ as a transformer network~\cite{Vaswani17}, due to its superior performance on time series tasks. Landmark coordinates for each time step are projected into tokens $\mathbf{x}_{\mathcal{E}} \in \mathbb{R}^{T \times d}$, and a masked attention mechanism is used to improve the temporal consistency of the output sequence.

The reconstruction loss is defined as the squared euclidean distance between the input $L$ and reconstructed $\hat{L} = f_{\theta}(L)$ landmark sequences:
\begin{equation}
    \mathcal{L}_{\mathit{rec}} = \frac{1}{T \cdot N_{\mathit{lnd}}} \sum_{i=1}^{T} \sum_{j=1}^{N_{\mathit{lnd}}}\lambda_j \lVert \hat{L}^i(j) - L^i(j) \rVert_2^2,
    \label{eq:loss-rec}
\end{equation}
where $\lambda_j$ denotes the weight of the j-th facial landmark in the loss calculation. We assign a higher weight value to non-rigid facial regions (i.e., eyes and mouth). 

Since the landmark configuration between consecutive time steps in real videos does not change significantly, we want the weights predicted by the encoder to also reflect this fact. For this, we add a regularization term to the loss function to minimize the distance between weights for consecutive time steps:
\begin{equation}
    \mathcal{L}_{\mathit{reg}} = \frac{1}{(T - 1) \cdot k} \sum_{i=1}^{T-1} \sum_{j=1}^{k} \lVert W^{i+1}(j) - W^i(j) \rVert_2^2.
    \label{eq:loss-reg}
\end{equation}

The final loss function is the weighted sum of both components: $\mathcal{L} = \mathcal{L}_{\mathit{rec}} + \lambda_{\mathit{reg}} \cdot \mathcal{L}_{\mathit{reg}}$, where $\lambda_{\mathit{reg}}$ is the weight of the regularization term. For our experiments, we set $\lambda_{\mathit{reg}} = 0.01$.

\begin{figure*}[!t]
    \centering
    \includegraphics[width=0.8\textwidth]{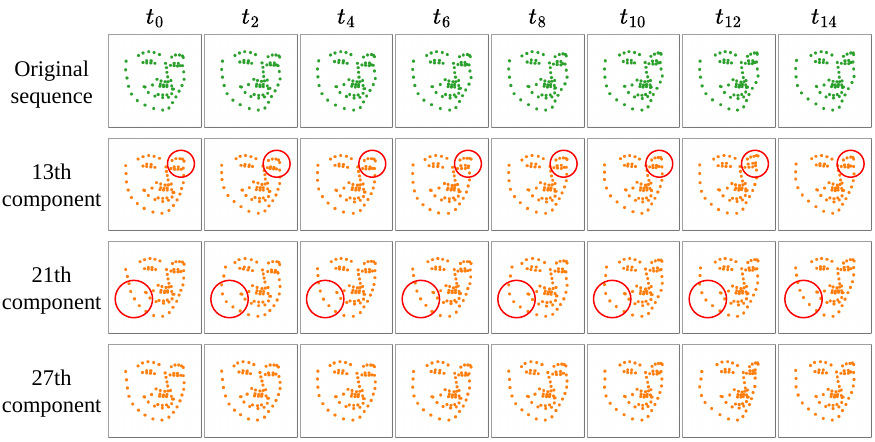}
    \caption{Examples of deformation bases learned by the LPN. Altering the weights estimated by the encoder for a particular basis causes the reconstructed landmark sequence to vary in different aspects. For visualization purposes, we only show one of every two time steps.}
    \label{fig:deformation-bases}
\end{figure*}

We use the CelebV-HQ dataset~\cite{Zhu22} to train the network, where we extract landmark annotations with the SPIGA detector~\cite{Prados-Torreblanca22}, for a total of $34,060$ landmark sequences. Once trained, we can generate a wide variety of temporal artifacts by introducing subtle noise to the weights predicted by the encoder for a particular deformation basis. For this, we randomly choose one column in $W$ following a uniform distribution $U\left\{1, \dots, k\right\}$, and add Gaussian noise $\mathcal{N}(0, \sigma^2)$, sampled independently for each time step, to the values of the chosen column. This procedure introduces slight variations to a particular deformation basis, which modifies one aspect of the reconstructed face in the temporal direction (see Fig.~\ref{fig:deformation-bases}). It is possible to generate more subtle or prominent artifacts by adjusting the noise variance $\sigma^2$. Formally, the manipulation process of a pristine landmark sequence $L_s$ can be defined as:
\begin{equation}
    L_t = \mathcal{D} \left( \mathcal{E} \left( L_s \right) + \mathcal{N} \left( 0, \sigma^2 \right) \right)
    \label{eq:manipulation-process}
\end{equation}

Furthermore, as shown in Fig.~\ref{fig:method}, we also perform a random sampling of face components to generate manipulated sequences. First, we sample a random number of face regions from the inner part of the face (i.e., eyebrows, eyes, nose, mouth), and only modify the landmarks from the selected regions, keeping the rest the same as the original sequence. This procedure is independent of the perturbations generated by the network and further improves the variety of samples generated by our method.

\noindent \textbf{Guided sampling of training data.} We have observed that, to effectively train the network to model rapid movements (e.g., blinks, quick mouth movements, etc.), we must pay special attention to the sampling procedure of training data. Since the temporal window of this type of movements is very small, if we naively sample random clips of length $T$ from landmark sequences following a uniform distribution, the network may be prone to ignore these movements, as they are statistically rare. To solve this, for each sequence, we sample clips following a Gaussian distribution centered around the time step $t$ with the largest non-rigid movements:
\begin{equation}
    \max_t \sum_{j \in \mathcal{J}} \lVert \delta L^t(j) \rVert_1, \ \text{such that} \ \mathcal{J} = \{ \mathit{eyes}, \mathit{mouth} \}.
    \label{eq:max-nonrigid}
\end{equation}

%-------------------------------------------------------------------------
\subsection{Face morphing}

After obtaining a sequence of landmarks generated by the LPN, our next goal is to introduce the underlying temporal artifacts into the video frames. To do so, we leverage a face morphing algorithm that modulates facial regions to match their corresponding positions in the target landmark sequence (see Algorithm~\ref{alg:face-morphing}).

\begin{algorithm}[H]
    \caption{Face morphing algorithm.}
    \label{alg:face-morphing}
    \begin{algorithmic}
        \STATE {\textbf{Input:} Source frames $I_s$, source landmarks $L_s$, target landmarks $L_t$, Delaunay triangulation $\mathcal{F}$}
        \STATE $I_t \gets I_s$
        \FOR{$i = 1, \dots, T$}
            \FORALL{$f \in \mathcal{F}$}
            \STATE $\mathbf{x}_s \gets \left\{L_s^i(j) \mid j \in f\right\}$
            \STATE $\mathbf{x}_t \gets \left\{L_t^i(j) \mid j \in f\right\}$
            \STATE $I^\prime_f \gets \mathit{Affine}(I_s^i, \mathbf{x}_s, \mathbf{x}_t)$
            \STATE $M_f \gets \mathit{MaskExtraction}(\mathbf{x}_t)$
            \STATE $I_t^i \gets I_t^i \odot (1 - M_f) + I^\prime_f \odot M_f$
            \ENDFOR
        \ENDFOR
        \STATE \textbf{return} $I_t$
    \end{algorithmic}
\end{algorithm}

In particular, the input consists of a sequence of source frames $I_s \in \mathbb{R}^{T \times H \times W \times C}$, source landmarks $L_s \in \mathbb{R}^{T \times N_{\mathit{lnd}} \times 2}$ and target landmarks $L_t \in \mathbb{R}^{T \times N_{\mathit{lnd}} \times 2}$, and the output is the modified frame sequence $I_t \in \mathbb{R}^{T \times H \times W \times C}$, initially set as a copy of $I_s$. Since the target landmark sequence $L_t$ is generated per-landmark, we perform triangulation-based morphing, in a manner similar to Active Appearance Models~\cite{Cootes01}. This requires a common definition of $N_{\mathit{tri}}$ triangles obtained with a Delaunay triangulation $\mathcal{F}= \left\{ (l, m, n)_g \mid l, m, n \in \mathbb{N}, 1 \leq l,m,n \leq N_{\mathit{lnd}} \right\}_{g=1}^{N_{\mathit{tri}}}$,  which contains landmark index triplets. This operation is performed only once to the canonical 68 landmark definition before training.

Then, for each frame in the sequence and each index triplet $f \in \mathcal{F}$, we first calculate the affine transformation that converts the pixel coordinates of the triangle defined in the source landmarks, $\mathbf{x}_s \in \mathbb{R}^{3 \times 2}$, to its corresponding triangle in the target landmarks, $\mathbf{x}_t \in \mathbb{R}^{3 \times 2}$, and apply this transformation to the input frame:
\begin{equation}
    I^\prime_f = \mathit{Affine}(I_s^i, \mathbf{x}_s, \mathbf{x}_t),    
    \label{eq:affine}
\end{equation}
where $I_s^i$ is the i-th frame in $I_s$ and $\mathbf{x}_\alpha = \left\{ L_\alpha^i(j) \mid j \in f\right\}$, $\alpha \in \left\{ s, t \right\}$. Finally, we generate a blending mask $M_f$ defined by the coordinates of the target triangle $\mathbf{x}_t$ from the landmark indices in $f$, and update the target frame with a blending operation:
\begin{equation}
    I_t^i = I_t^i \odot (1 - \sum_{f \in \mathcal{F}} M_f) + \sum_{f \in \mathcal{F}} (I^\prime_f \odot M_f).
    \label{eq:blend}
\end{equation}

Algorithm~\ref{alg:face-morphing} computes Eq.~\ref{eq:blend} but updates $I_t^i$ with each triangle $f$ iteratively.

The final frame sequence $I_t$ contains temporal artifacts introduced by the LPN, which are complementary to the spatial artifacts introduced by SBI~\cite{Shiohara22}, when uniformly applied to all frames in a video clip with the same parameters (i.e., $\sum \lVert \Delta_t \rVert_1 = 0$, as per Eq.~\ref{eq:temporal-artifacts}). For this reason, we can combine the two pseudo-fake generation techniques to train a network to learn both spatial and temporal clues. Furthermore, training the network with samples containing pure spatial or temporal artifacts (as opposed to mixing the two categories in the same video clip), alleviates the complexity imbalance of temporal clues during training~\cite{Zheng21, Wang23a}, as spatial clues generated by face swap are usually more dominant in deepfake videos, and the network tends to ignore the more complex temporal clues.

%-------------------------------------------------------------------------
\section{Experiments}

\begin{table*}[!t]
    \caption{Comparison with SOTA Methods in Terms of Video-level AUC (\%). ``Pristine Only'' Refers to Methods That Train Exclusively with Pseudo-fakes. Results Marked with $\dagger$ Are Computed by Us with Inference Code and Model Weights Provided by Authors}
    \label{tab:cross-dataset-eval}
    \centering
    \scriptsize
    \begin{tabular}{@{}l@{}ccccccccccc@{}}
        \toprule
        \multirow{2}{*}{\textbf{Method}} & \multirow{2}{*}{\textbf{Pristine only}} & \multicolumn{6}{c}{\textbf{Cross-dataset}} & \multicolumn{4}{c}{\textbf{DF40}} \\
        \cmidrule(lr){3-8} \cmidrule(l){9-12}
        & & \textbf{CDF} & \textbf{DFD} & \textbf{DFDCP} & \textbf{WDF} & \textbf{DFo} & \textbf{Avg.} & \textbf{BlendFace} & \textbf{FSGAN} & \textbf{MobileSwap} & \textbf{Avg.} \\
        \midrule
        F3Net~\cite{Qian20} & \xmark & 78.90 & 84.40 & 74.90 & 72.80 & 73.00 & 76.80 & 80.80 & 84.50 & 86.70 & 84.00 \\
        SPSL~\cite{Liu21} & \xmark & 79.90 & 87.10 & 77.00 & 70.20 & 72.30 & 77.30 & 74.80 & 81.20 & 88.50 & 81.50 \\
        RECCE~\cite{Cao22} & \xmark & 82.30 & 89.10 & 73.40 & 75.60 & 78.40 & 79.76 & 83.20 & 94.90 & 92.50 & 90.20 \\
        SLADD~\cite{Chen22} & \xmark & 83.70 & 90.40 & 75.60 & 69.00 & 80.00 & 79.74 & 88.20 & 94.30 & 95.40 & 92.63 \\
        UCF~\cite{Yan23b} & \xmark & 83.70 & 86.70 & 77.00 & 77.40 & 80.80 & 81.12 & 82.70 & 93.70 & 95.00 & 90.47 \\
        SRM~\cite{Luo21} & \xmark & 84.00 & 88.50 & 72.80 & 70.20 & 72.20 & 77.54 & 70.40 & 77.20 & 77.90 & 75.17 \\
        MSVT~\cite{Yu23} & \xmark & 88.81 & 91.36 & - & - & 98.42 & - & - & - & - & - \\
        Altfreezing~\cite{Wang23a} & \xmark & 89.50 & 98.50 & 70.91 & \quad 59.58~$\dagger$ & 99.30 & 83.56 & 95.10 & 95.80 & 85.10 & 92.00 \\
        SFA~\cite{Cobo25} & \xmark & 89.52 & \quad 86.74~$\dagger$ & 80.58 & \quad 73.70~$\dagger$ & 99.24 & 85.96 & - & - & - & - \\
        PTF~\cite{Kim25} & \xmark & 89.70 & 97.30 & - & - & 99.40 & - & - & - & - & - \\
        FSFM~\cite{Wang25} & \xmark & 91.44 & - & 89.71 & 86.96 & - & - & - & - & - & - \\
        FakeRadar~\cite{Li25a} & \xmark & 91.70 & 96.20 & 88.50 & - & - & - & - & - & - & - \\
        CDFA~\cite{Lin24} & \xmark & 93.80 & 95.40 & 88.10 & 79.60 & 97.30 & 90.84 & 75.60 & 94.20 & 82.30 & 84.03 \\
        FCG~\cite{Han25} & \xmark & \textbf{95.00} & \quad 93.00~$\dagger$ & \quad 86.59~$\dagger$ & \textbf{87.20} & \textbf{99.60} & 92.28 & - & - & - & - \\
        SeeABLE~\cite{Larue23} & \cmark & 87.30 & - & 86.30 & - & - & -  & - & - & - & - \\
        DeepShield~\cite{Cai25} & \cmark & 92.20 & 96.10 & 93.20 & - & - & -  & - & - & - & - \\
        FakeSTormer~\cite{Nguyen25} & \cmark & 92.80 & \textbf{98.60} & 90.20 & 75.30 & - & - & 91.10 & 96.40 & 95.00 & 94.17 \\
        SBI (c23)~\cite{Shiohara22} & \cmark & 92.87 & 98.16 & 85.51 & \quad 73.56~$\dagger$ & \quad 84.36~$\dagger$ & 86.90 & 89.10 & 80.30 & 95.20 & 88.20 \\
        \midrule
        LSDA~\cite{Yan24}~$*$ & \xmark & 89.80 & 95.60 & 81.20 & 75.60 & 89.20 & 86.28 & 87.50 & 93.90 & 93.00 & 91.47 \\
        Effort~\cite{Yan25a}~$*$ & \xmark & 95.60 & 96.50 & 90.90 & 84.80 & 97.70 & 93.10 & 87.30 & 95.70 & 95.30 & 92.77 \\
        VB~\cite{Yan25b}~$*$ & \cmark & 94.70 & 96.50 & 90.90 & 84.80 & 99.10 & 93.20 & 90.60 & 96.40 & 94.60 & 93.87 \\
        \midrule
        \textbf{\ourmethod{} (ours)} & \cmark & 94.74 & 97.04 & \textbf{93.89} & 83.52 & 98.47 & \textbf{93.53} & \textbf{98.06} & \textbf{98.05} & \textbf{96.86} & \textbf{97.66} \\
        \bottomrule
    \end{tabular}
\end{table*}

\noindent \textbf{Datasets}. Following the standard practice in the literature, we use FaceForensics++ (FF++)~\cite{Rossler19} as the training dataset. It consists of 1,000 real and 4,000 fake videos constructed using 4 manipulation methods. Similar to~\cite{Li20a, Shiohara22, Larue23, Nguyen25}, we discard the original fake videos and use only pseudo-fake samples to train the network. The dataset provides 3 compression quality settings: raw, light compression (c23) and heavy compression (c40). As established by the literature, we employ the c23 subset, also denoted as HQ (high-quality) on other papers.

For cross-dataset evaluation, we use 5 datasets: Celeb-DFv2 (CDF)~\cite{Li20b} (518 test videos), DFD~\cite{DFD} (3,431 test videos), DFDCP~\cite{Dolhansky19} (780 test videos), WildDeepFake (WDF)~\cite{Zi20} (806 test videos) and DeeperForensics-1.0 (DFo)~\cite{Jiang20} (280 test videos). For further comparison, we also employ the recently released DF40 dataset~\cite{Yan23a}, which contains videos from the test subset of FF++ manipulated with state-of-the-art (SOTA) deepfake generation methods. Following~\cite{Nguyen25}, we use the BlendFace, FSGAN and MobileSwap subsets.

\noindent \textbf{Evaluation metrics}. To compare our method with other SOTA deepfake detectors, we report video-level Area Under the Receiver-operating Characteristic Curve (AUC), by averaging the predictions of the network over all non-overlapping clips in a video sequence. We also provide Equal Error Rate (EER) metrics in Tab.~\ref{tab:backbone-ablation} for further understanding of our results.

\noindent \textbf{Implementation details}. We use RetinaFace~\cite{Deng20} to extract bounding boxes and SPIGA~\cite{Prados-Torreblanca22} for $N_{\mathit{lnd}}=68$ landmark annotations. We also align the center of the bounding box with the tip of the nose for each frame. As the deepfake detection network, we fine-tune a pretrained MARLIN encoder~\cite{Cai23}, adding an extra classification token for the deepfake detection task. The network receives sequences of $T=16$ consecutive $224 \times 224 \times 3$ frames and outputs classification logits, which are compared to the ground-truth labels with a binary cross-entropy loss function. We follow \cite{Shiohara22} to construct batches containing real and pseudo-fake samples, but randomly choose either spatial or temporal pseudo-fakes for each sample. The data augmentation pipeline includes clip-level horizontal flip, chromatic manipulations, gaussian blur, JPEG compression, pixel dropout, affine transformations and random cropping. We use Adam optimization~\cite{Kingma15} to train the network for around 300 epochs with a maximum learning rate of $7 \cdot 10^{-6}$ and a cosine annealing scheduler with linear warm-up for the first 30 epochs. We select the model checkpoint with the lowest average loss on the validation subset of FF++, consisting of real videos and the same pseudo-fake generation methods used in the train subset. For the LPN, we set $\sigma=0.007$ as the standard deviation of the Gaussian noise, and $k=64$ deformation bases, following an ablation study.

%-------------------------------------------------------------------------
\subsection{Cross-dataset evaluation}

Table~\ref{tab:cross-dataset-eval} shows a comparison of our approach with other SOTA deepfake detection methods. Our results correspond to a ViT-L MARLIN encoder trained with a combination of spatial and our proposed temporal pseudo-fakes. As we can see, our method achieves the highest average AUC, and sets a new SOTA result on DFDCP. On DF40, we achieve SOTA results on all subsets, yielding an average improvement of more than 3 points and demonstrating the generalization capabilities of our approach to a diverse range of SOTA deepfake generation techniques. The lowest AUC result for our model corresponds to the WDF dataset~\cite{Zi20}. Note that this dataset provides pre-processed face crops instead of raw videos, which results in an extra performance drop, as our pre-processing pipeline is different. When following the pre-processing of the WDF dataset, we achieve an AUC of 86.03\%. Nevertheless, our network demonstrates competitive performance even under different pre-processing variations, and surpasses methods that require deepfake videos from FF++ in combination with pseudo-fake samples~\cite{Chen22, Wang23a, Lin24}. It is also worth noting that we achieve the best generalization results without using any special network architecture or complicated multi-task framework optimized for deepfake detection, as other methods~\cite{Han25, Nguyen25, Yan25b}. Existing approaches in the literature can be easily combined with our training data to achieve further improvements.

Note also that some methods~\cite{Yan24, Yan25a, Yan25b}, marked with "*" in Table~\ref{tab:cross-dataset-eval}, use metrics computed from test datasets to select the best performing model following the \textit{DeepfakeBench} codebase\footnote{\url{https://github.com/SCLBD/DeepfakeBench}}, which yields over-optimistic results. Still, we also surpass their results despite using videos from the FF++ validation subset to perform the model selection.

%-------------------------------------------------------------------------
\subsection{Ablation analysis}

\begin{figure*}[!t]
    \centering
    \includegraphics[width=0.99\textwidth]{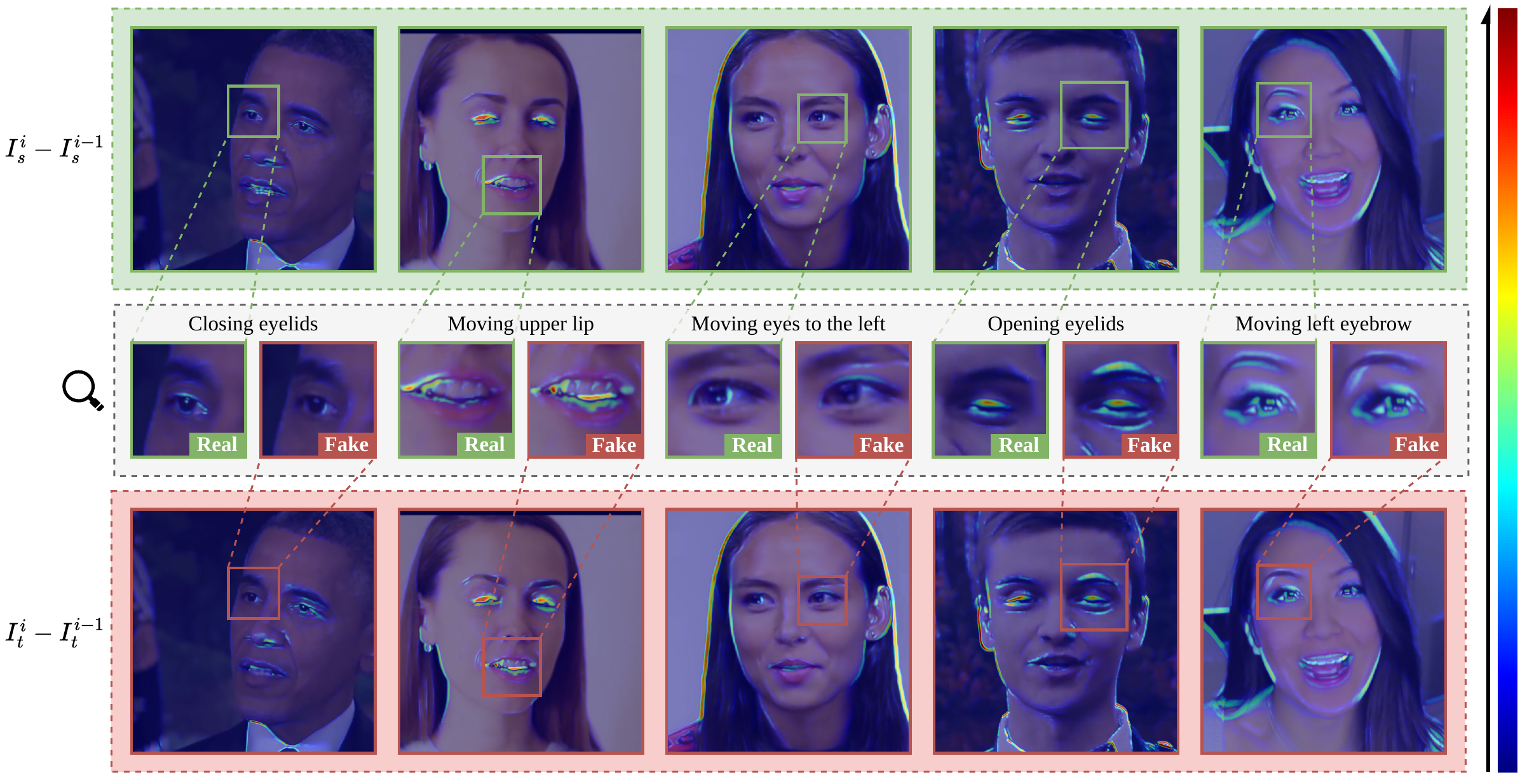}
    \caption{Visualization of subtle temporal artifacts introduced by our method (bottom row) compared to the facial movement of the corresponding real sequence (top row), extracted from pristine videos of the FF++ dataset~\cite{Rossler19}.}
    \label{fig:ours-examples}
\end{figure*}

Here we report the results obtained with the ViT-B configuration for the MARLIN encoder, unless otherwise specified.

\begin{table}[!t]
    \caption{Video-level AUC (\%) Obtained with Different Combinations of Pseudo-fake Generation Methods. First Row Shows the Results of a Baseline Model Trained with Original Fakes from FF++~\cite{Rossler19}.}
    \label{tab:pseudo-fake-ablation}
    \centering
    \scriptsize
    \begin{tabular}{@{}cccccccc@{}}
        \toprule
        \textbf{Spatial} & \textbf{Temporal} & \textbf{CDF} & \textbf{DFD} & \textbf{DFDCP} & \textbf{WDF} & \textbf{DFo} & \textbf{Avg.} \\
        \midrule
        \xmark & \xmark & 85.16 & 87.70 & 79.44 & 72.63 & \textbf{99.18} & 84.82 \\
        \cmark & \xmark & \textbf{93.75} & 87.32 & \textbf{91.34} & 81.76 & 75.12 & 85.86 \\
        \xmark & \cmark & 64.76 & \textbf{99.13} & 76.63 & 72.15 & 98.21 & 82.18 \\
        \cmark & \cmark & 90.10 & 97.05 & 90.84 & \textbf{81.80} & 97.79 & \textbf{91.52} \\
        \bottomrule
    \end{tabular}
\end{table}

\noindent \textbf{Combination of spatial and temporal pseudo-fakes}. Table~\ref{tab:pseudo-fake-ablation} shows the results of the network trained with different combinations of pseudo-fake generation techniques. The first row shows the results of a baseline model trained on the original fake videos from FF++. Then, we can see how a model trained on individual pseudo-fake generation techniques (i.e., spatial or temporal) only performs adequately on a subset of testing datasets, but achieves a worse average result compared to the baseline, as the network is only aware of a particular manipulation modality. However, when we combine spatial and temporal pseudo-fakes, we obtain the model with the best generalization performance, increasing the average AUC score from the baseline model by more than 6 points.

These results also showcase the different degrees of spatial and temporal clues introduced by each dataset. For example, we can see how fake videos in CDF or DFDCP can be easily identified by their spatial artifacts, but videos in DFD or DFo are more dominated by temporal artifacts. In particular, we achieve a SOTA result of 99.13\% AUC on DFD when training only with our proposed temporal pseudo-fake samples.

\begin{table}[!t]
    \caption{Comparison Between Different Approaches for Introducing Temporal Artifacts. All Networks Are Trained with a Combination of Spatial and Temporal Artifacts.}
    \label{tab:noise-ablation}
    \centering
    \scriptsize
    \begin{tabular}{@{}lcccccc@{}}
        \toprule
        \textbf{Temporal artifacts} & \textbf{CDF} & \textbf{DFD} & \textbf{DFDCP} & \textbf{WDF} & \textbf{DFo} & \textbf{Avg.} \\
        \midrule
        Spatial baseline & 93.75 & 87.32 & \textbf{91.34} & 81.76 & 75.12 & 85.77 \\
        Noise (iid) & 93.19 & 91.04 & 88.21 & 81.18 & 88.43 & 88.41 \\
        Noise (multivariate) & \textbf{93.98} & 90.61 & 90.88 & \textbf{83.16} & 87.46 & 89.22 \\
        LPN & 90.10 & \textbf{97.05} & 90.84 & 81.80 & \textbf{97.79} & \textbf{91.52} \\
        \bottomrule
    \end{tabular}
\end{table}

\begin{figure*}[!t]
    \centering
    \subfloat[]{\includegraphics[width=0.33\textwidth]{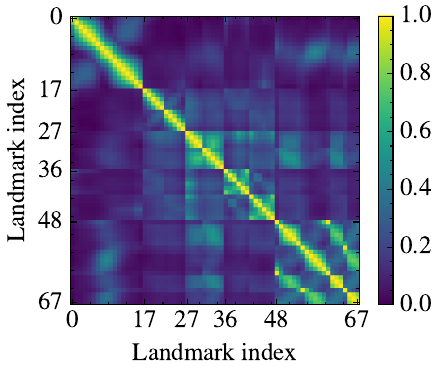}%
    \label{fig:dfo-correlation}}
    \hfill
    \subfloat[]{\includegraphics[width=0.33\textwidth]{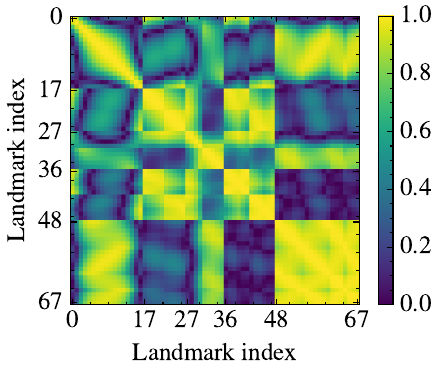}%
    \label{fig:vb-corrrelation}}
    \hfill
    \subfloat[]{\includegraphics[width=0.33\textwidth]{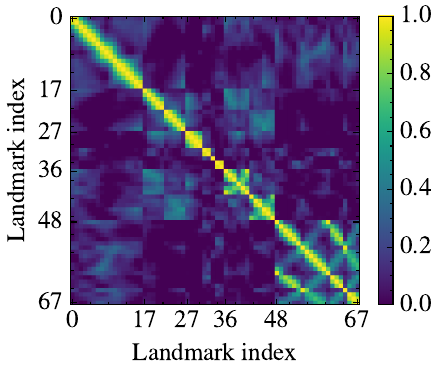}%
    \label{fig:ours-correlation}}
    \caption{Correlation matrices of temporal artifacts (Eq.~\ref{eq:temporal-artifacts}) computed from landmark motions extracted from deepfake videos~\cite{Jiang20} (a), VB~\cite{Yan25b} (b) and our method (c). Landmark indices correspond to the common definition of Multi-PIE~\cite{Gross08}.}
    \label{fig:temporal-artifacts-correlation}
\end{figure*}

\noindent \textbf{Comparison between different methods for introducing temporal artifacts.} To validate the importance of the LPN, we test other analytical approaches by directly manipulating landmark coordinates with subtle Gaussian noise. We designed two variants: noise sampled independently for each landmark, and multivariate noise that considers the correlation between different facial regions in temporal artifacts caused by deepfake videos (as shown in Fig.~\ref{fig:dfo-correlation}). The manipulated sequence then undergoes the region sampling and face morphing modules depicted in Fig.~\ref{fig:method}. Table~\ref{tab:noise-ablation} shows the results of the network trained on a combination of spatial artifacts and this modified version of our method.

As the results show, the best generalization performance is achieved when we use the LPN, demonstrating that the artifacts introduced by the network are more semantically rich and essential to reliably identify deepfake videos. As shown in Tab.~\ref{tab:pseudo-fake-ablation}, DFD and DFo are the datasets that better represent the importance of temporal pseudo-fakes, as spatial clues are not as useful as in other datasets for our video-level detector. When we use our network to generate temporal artifacts, we improve the AUC results by 7 points on DFD and 9 points on DFo compared to the alternative simple and multivariate noise-based approaches.

Furthermore, we show several examples of the temporal artifacts introduced by our method in Fig.~\ref{fig:ours-examples}. Our method is able to generate new facial motions not present in the original videos, as well as introduce kinematic inconsistencies in a manner similar to the example shown in Fig.~\ref{fig:deepfake-artifacts}, as exemplified in the first and third columns of Fig.~\ref{fig:ours-examples}, where the motion present in the original sequences is removed. To the best of our knowledge, this phenomenon cannot be reproduced by any approach in the current literature.

\begin{table}[!t]
    \caption{Comparison Between Different Temporal Pseudo-fake Generation Methods. All Networks Are Trained with a Combination of Spatial and the Corresponding Temporal Pseudo-fakes. Note That Neither~\cite{Wang23a, Yan25b} Provide Code, So We Tested Our Implementation.}
    \label{tab:temporal-artifacts-ablation}
    \centering
    \addtolength{\tabcolsep}{-0.2em}
    \begin{tabular}{@{}lcccccc@{}}
        \toprule
        \textbf{Method} & \textbf{CDF} & \textbf{DFD} & \textbf{DFDCP} & \textbf{WDF} & \textbf{DFo} & \textbf{Avg.} \\
        \midrule
        CBI (MARLIN ViT-B) & \textbf{94.68} & 86.66 & 89.75 & 79.17 & 76.18 & 85.29 \\
        VB (MARLIN ViT-B) & 91.80 & 95.18 & 87.37 & \textbf{82.42} & 95.41 & 90.44 \\
        \ourmethod{} (ours) & 90.10 & \textbf{97.05} & \textbf{90.84} & 81.80 & \textbf{97.79} & \textbf{91.52} \\
        \bottomrule
    \end{tabular}
\end{table}

\noindent \textbf{Comparison with other video-level pseudo-fake generation methods.} Table~\ref{tab:temporal-artifacts-ablation} shows a comparison of our method with other SOTA pseudo-fake video generators, in combination with spatial artifacts. On average, our method is able to introduce more generalizable temporal artifacts. 

CBI~\cite{Wang23a} is prone to introduce unrealistic temporal artifacts, yielding poor results on datasets oriented towards this type of manipulations (e.g., DFD and DFo). VB~\cite{Yan25b} performs random affine transformations to the inner part of the face for each frame independently, which is a better approximation than the previous method, but still is not able to accurately model complex temporal artifacts, as demonstrated by the results on DFD and DFo, where our method outperforms VB by approximately 2 points.

Our data-driven approach achieves the best generalization performance and provides a flexible framework to improve the quality of pseudo-fake videos used to train the network. Furthermore, Fig.~\ref{fig:temporal-artifacts-correlation} shows the correlation matrices of the temporal artifacts generated by VB and our approach, compared to the correlation found on actual deepfake videos. As the figures illustrate, the correlations produced by VB exhibit a highly structured pattern. This is a direct consequence of its use of holistic affine transformations, which model facial regions as rigid units. In contrast, our approach generates a correlation matrix that more closely resembles the one produced by genuine deepfakes.

\begin{table}[!t]
    \caption{Video-level AUC$\uparrow$ (\%, top) and EER$\downarrow$ (\%, bottom) Results for Different Backbone Configurations of the MARLIN Encoder~\cite{Cai23}.}
    \label{tab:backbone-ablation}
    \centering
    \scriptsize
    \setlength{\tabcolsep}{0.7em}
    \begin{tabular}{@{}lccccccc@{}}
        \toprule
        \textbf{Backbone} & \textbf{Parameters} & \textbf{CDF} & \textbf{DFD} & \textbf{DFDCP} & \textbf{WDF} & \textbf{DFo} & \textbf{Avg.} \\
        \midrule
        ViT-B & 86.2M & 90.10 & \textbf{97.05} & 90.84 & 81.80 & 97.79 & 91.52 \\
        ViT-L & 303.9M & \textbf{94.74} & 97.04 & \textbf{93.89} & \textbf{83.52} & \textbf{98.47} & \textbf{93.53} \\
        \midrule
        ViT-B & 86.2M & 17.75 & 9.76 & 18.88 & 27.17 & 6.43 & 16.00 \\
        ViT-L & 303.9M & \textbf{11.40} & \textbf{7.97} & \textbf{13.73} & \textbf{25.93} & \textbf{2.14} & \textbf{12.23} \\
        \bottomrule
    \end{tabular}
\end{table}

\noindent \textbf{Comparison between different model sizes}. To demonstrate the scalability of our method, we train medium and large-sized versions of the MARLIN encoder, and obtain a performance improvement with the larger model across all test datasets, despite the limited size of the training dataset and the lack of regularization techniques other than data augmentation (see Tab.~\ref{tab:backbone-ablation}). As our approach only needs real videos to train, gathering large amounts of data to train a deepfake detection model for its use in real world scenarios becomes significantly easier, and the generic artifacts introduced by our method make the detector more resistant to overfitting to particular deepfake generation techniques.

\noindent \textbf{LPN regularization loss.}
Table~\ref{tab:reg-loss} shows the performance gains of the regularization loss included in the training of the LPN (Eq.~\ref{eq:loss-reg}). As the results show, when we remove this term from the loss calculation, the resulting network is unable to produce temporal artifacts of the same quality as the proposed approach. Specifically, adding the regularization term increases the overall video-level AUC by 2 points, demonstrating the effectiveness of this loss term to produce more useful temporal artifacts.

\begin{table}[!t]
    \caption{Deepfake Detection Results Obtained By Including or Excluding the Regularization Loss Term for Training the LPN.}
    \label{tab:reg-loss}
    \centering
    \begin{tabular}{@{}ccccccc@{}}
        \toprule
        $\mathcal{L}_{\mathit{reg}}$ & \textbf{CDF} & \textbf{DFD} & \textbf{DFDCP} & \textbf{WDF} & \textbf{DFo} & \textbf{Avg.} \\
        \midrule
        \xmark & 86.62 & 94.94 & \textbf{92.73} & 80.13 & 93.37 & 89.56 \\
        \cmark & \textbf{90.10} & \textbf{97.05} & 90.84 & \textbf{81.80} & \textbf{97.79} & \textbf{91.52} \\
        \bottomrule
    \end{tabular}
\end{table}

\noindent \textbf{Region sampling module.}
Table~\ref{tab:region-sampling} shows an extension of Table~\ref{tab:noise-ablation}, where we demonstrate the effectiveness of the region sampling module shown on Fig.~\ref{fig:method}. Here we compare results obtained with iid Gaussian noise and the proposed LPN, both with and without the face region sampling module. As the results show, the region sampling module improves the results by 4 points in the case of Gaussian noise and more than 5 points when using the LPN when compared to the spatial baseline. The combination of LPN and region sampling improves by 0.62 points the bare LPN. This indicates that this module generates a richer variety of temporal artifacts, and the disruption of correlated movements in the face yields generic clues more aligned to real deepfake videos, even under simple analytical approximations.

\begin{table}[!t]
    \caption{Deepfake Detection Results Obtained By Including or Excluding the Region Sampling Module in the Proposed Pseudo-fake Generation Pipeline.}
    \label{tab:region-sampling}
    \centering
    \addtolength{\tabcolsep}{-0.3em}
    \begin{tabular}{@{}lccccccc@{}}
        \toprule
        \multirow{2}{*}{\textbf{Method}} & \textbf{Region} & \multirow{2}{*}{\textbf{CDF}} & \multirow{2}{*}{\textbf{DFD}} & \multirow{2}{*}{\textbf{DFDCP}} & \multirow{2}{*}{\textbf{WDF}} & \multirow{2}{*}{\textbf{DFo}} & \multirow{2}{*}{\textbf{Avg.}} \\
        & \textbf{sampling} & & & & & & \\
        \midrule
        Spatial baseline & \xmark & \textbf{93.75} & 87.32 & \textbf{91.34} & \textbf{81.76} & 75.12 & 85.77 \\
        Noise (iid) & \xmark & 86.18 & 89.80 & 91.05 & 80.26 & 82.47 & 84.22 \\
        Noise (iid) & \cmark & 93.19 & 91.04 & 88.21 & 81.18 & 88.43 & 88.41 \\
        LPN & \xmark & 92.67 & 92.80 & 90.84 & 81.57 & 96.62 & 90.90 \\
        \textbf{LPN} & \cmark & 90.10 & \textbf{97.05} & 90.84 & 81.80 & \textbf{97.79} & \textbf{91.52} \\
        \bottomrule
    \end{tabular}
\end{table}

\noindent \textbf{LPN hyperparameters.}
In this study, we test the ViT-B detector with different hyperparameters of the LPN. Specifically, we tested several values for the standard deviation of the gaussian noise used to introduce temporal artifacts $\sigma$ and the number of deformation bases $k$. Table~\ref{tab:lpn-hyperpatameters} shows the results of deepfake detectors trained on pseudo-fakes with different combinations for these hyperparameters.

\begin{table}[!t]
    \caption{Results for Different Combinations of LPN Hyperparameters.}
    \label{tab:lpn-hyperpatameters}
    \centering
    \begin{tabular}{@{}cccccccc@{}}
        \toprule
        $\sigma$ & $k$ & \textbf{CDF} & \textbf{DFD} & \textbf{DFDCP} & \textbf{WDF} & \textbf{DFo} & \textbf{Avg.} \\
        \midrule
        0.01 & 64 & 90.20 & 96.03 & 89.23 & 81.39 & 97.70 & 90.91 \\
        0.007 & 64 & 90.10 & \textbf{97.05} & 90.84 & 81.80 & 97.79 & \textbf{91.52} \\
        0.005 & 64 & 89.15 & 96.01 & 90.64 & \textbf{82.82} & 97.49 & 91.22 \\
        0.001 & 64 & 87.14 & 90.79 & \textbf{91.08} & 81.07 & 85.94 & 87.20 \\
        0.007 & 32 & \textbf{91.06} & 95.78 & 88.57 & 82.31 & \textbf{97.81} & 91.11 \\
        \bottomrule
    \end{tabular}
\end{table}

\noindent \textbf{Face reenactment results.}
Since our work focuses on inconsistencies between facial movements, we also want to provide results on face reenactment manipulation benchmarks, where the original identity of the subject is maintained but the expression, such as the movement of the mouth, is generated to mimic that of another video. Table~\ref{tab:wav2lip} shows the results of the detectors evaluated in the \textit{wav2lip} subset of the DF40 dataset~\cite{Yan23a}. We selected this subset because it is the only one that contains uncropped videos for the complete FF++ test set, which is necessary for this analysis. As we can see, our method outperforms CBI~\cite{Wang23a} and VB~\cite{Yan25b} by a significant margin in both metrics when using ViT-B and ViT-L detectors. This demonstrates the superiority of our approach in modeling complex temporal artifacts, as face reenactment methods tend to introduce less obvious spatial artifacts, and the clues located in the temporal dimension become more important for the detector to reliably identify manipulated content.

\begin{table}[!t]
    \caption{Results on the Wav2lip Subset of the DF40 Dataset~\cite{Yan23a}.}
    \label{tab:wav2lip}
    \centering
    \addtolength{\tabcolsep}{-0.1em}
    \begin{tabular}{@{}lcccc@{}}
        \toprule
        \multirow{2}{*}{\textbf{Method}} & \multicolumn{2}{c}{\textbf{ViT-B}} & \multicolumn{2}{c}{\textbf{ViT-L}} \\
        \cmidrule(lr){2-3} \cmidrule(l){4-5}
        & \textbf{AUC~$\uparrow$ (\%)} & \textbf{EER~$\downarrow$ (\%)} & \textbf{AUC~$\uparrow$ (\%)} & \textbf{EER~$\downarrow$ (\%)}\\
        \midrule
        CBI & 71.18 & 31.79 & 83.45 & 24.29 \\
        VB & 80.60 & 28.21 & 93.29 & 13.21 \\
        \textbf{KiMoI (ours)} & \textbf{82.37} & \textbf{24.64} & \textbf{96.35} & \textbf{10.00} \\
        \bottomrule
    \end{tabular}
\end{table}

\begin{figure}[!t]
    \centering
     \includegraphics[width=0.47\textwidth]{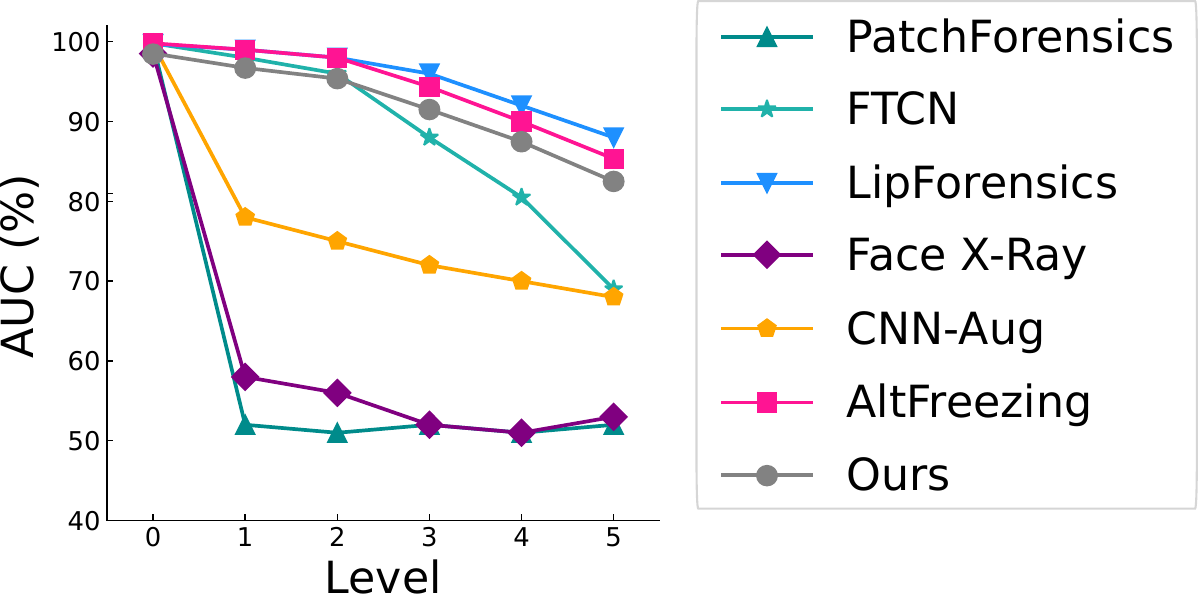}
    \caption{Video compression robustness evaluation. We employed videos from the DeeperForensics-1.0 benchmark~\cite{Jiang20} with different levels of video compression, which is not seen during training.}
    \label{fig:video-compresssion}
\end{figure}

\noindent \textbf{Robustness against video compression.} In this study, we test the robustness of our method against video compression, common in many online platforms such as social media. The compression algorithm reduces the amount of information between consecutive frames to generate a smaller video file, which obfuscates the process of identifying deepfake artifacts. As established in the literature, we employ the DeeperForensics-1.0~\cite{Jiang20} benchmark to evaluate the performance of our network compared to other methods in the literature. Following~\cite{Wang23a}, our comparison includes PatchForensics~\cite{Chai20}, FTCN~\cite{Zheng21}, LipForensics~\cite{Haliassos21}, Face X-Ray~\cite{Li20a}, CNN-Aug~\cite{Wang20a} and AltFreezing~\cite{Wang23a}. Fig.~\ref{fig:video-compresssion} shows the results with different levels of video compression. Our method achieves results similar to compression-resistant SOTA methods, such as LipForensics~\cite{Haliassos21} and AltFreezing~\cite{Wang23a}, demonstrating the effectiveness of our approach to perturbations commonly found in in-the-wild data.

%-------------------------------------------------------------------------
 \section{Conclusions}

In this paper, we presented a novel technique for generating pseudo-fake videos with subtle, generic temporal artifacts. This allowed us to train a deepfake video detection network that is not dependent on any original fake samples for its training. Empirically, it achieves a new state-of-the-art generalization across a combination of popular deepfake detection benchmarks. We demonstrated that our synthesis approach provides more robust and transferable temporal clues than comparable methods in the literature. 

The flexibility of our framework opens up future research avenues to learn interpretable deformation modes that produce semantically rich synthetic data.
Furthermore, by allowing for the targeted manipulation of specific facial regions during training, we may also significantly improve the interpretability of a detector's final decision.

Our results confirm that kinematics is a distinct and complementary signal. By proving that motion consistency is learnable from synthetic data, our work establishes a new cue for data-driven temporal fake detection, providing an essential ingredient for the next generation of generalizable face forgery detection methods.

Code will be released upon publication.

\bibliographystyle{IEEEtran}
\bibliography{main}

\end{document}